%% file: main.tex
\documentclass[10pt,twocolumn,letterpaper]{article}

\usepackage{wacv}              

\usepackage{graphicx}
\usepackage{amsmath}
\usepackage{amssymb}
\usepackage{booktabs}
\usepackage{siunitx}
\usepackage{balance}
\usepackage[accsupp]{axessibility}
 
\usepackage[pagebackref,breaklinks,colorlinks]{hyperref}

\usepackage[capitalize]{cleveref}
\crefname{section}{Sec.}{Secs.}
\Crefname{section}{Section}{Sections}
\Crefname{table}{Table}{Tables}
\crefname{table}{Tab.}{Tabs.}

\def\wacvPaperID{2} 
\def\confName{WACV}
\def\confYear{2024}

\begin{document}

\title{Forensic Iris Image Synthesis}

\author{Rasel Ahmed Bhuiyan\hspace{1cm}Adam Czajka\vspace{0.1cm}\\
Department of Computer Science and Engineering, University of Notre Dame\\
384 Fitzpatrick Hall of Engineering, Notre
Dame, 46556, Indiana, USA\\
{\tt\small \{rbhuiyan,aczajka\}@nd.edu}
}

\maketitle

\begin{abstract}

Post-mortem iris recognition is an emerging application of iris-based human identification in a forensic setup, able to correctly identify deceased subjects even three weeks post-mortem. This technique thus is considered as an important component of future forensic toolkits. The current advancements in this field are seriously slowed down by exceptionally difficult data collection, which can happen in mortuary conditions, at crime scenes, or in ``body farm'' facilities. This paper makes a novel contribution to facilitate progress in post-mortem iris recognition by offering a conditional StyleGAN-based iris synthesis model, trained on the largest-available dataset of post-mortem iris samples acquired from more than 350 subjects, generating -- through appropriate exploration of StyleGAN latent space -- multiple within-class (same identity) and between-class (different new identities) post-mortem iris images, compliant with ISO/IEC 29794-6, and with decomposition deformations controlled by the requested PMI (post mortem interval). Besides an obvious application to enhance the existing, very sparse, post-mortem iris datasets to advance -- among others -- iris presentation attack endeavors, we anticipate it may be useful to generate samples that would expose professional forensic human examiners to never-seen-before deformations for various PMIs, increasing their training effectiveness. The source codes and model weights are made available with the paper.

\end{abstract}

\input{figure-tex/teaser}

\section{Introduction}
\label{sec:intro}
\input{sections/introduction}

\section{Related Works}
\label{sec:related_works}
\input{sections/related-works}

\section{Methodology}
\label{sec:method}
\input{sections/methodology}

\input{figure-tex/class-wise-authentic-synthetic-samples}
\input{figure-tex/noise-level-sample}
\input{figure-tex/class-wise-score-dist}
\input{figure-tex/entire-data-iso-score-dist}

\section{Experimental Analysis}
\label{sec:exp}
\input{sections/experiment}

\section{Summary}
\label{sec:conclusion}
\input{sections/conclusion}

\balance

{\small
\bibliographystyle{ieee_fullname}
\bibliography{ref}
}

\clearpage

\appendix

\nobalance

\begin{center}
\large{\bf Forensic Iris Image Synthesis}\\
Supplementary materials: PMI-specific comparison score distributions
\end{center}

Fig. \ref{fig:class-wise-noise-level-score-dist} shows genuine score distributions obtained for authentic post-mortem samples and synthesized post-mortem samples for different sizes of the hypersphere (defined by radius $\varepsilon_{\max}$, shown in parentheses) defining the same-identity manifold in the StyleGAN's latent space $\mathbf{w}$, broken by the PMI range (StyleGAN condition). The plots suggest that setting different $\varepsilon_{\max}$ for different PMI ranges may be beneficial. 
\input{figure-tex/class-wise-noise-level-score-dist}

\end{document}

%% file: figure-tex/teaser.tex
\begin{figure*}[!ht] 
    \centering
    \includegraphics[width=\textwidth]{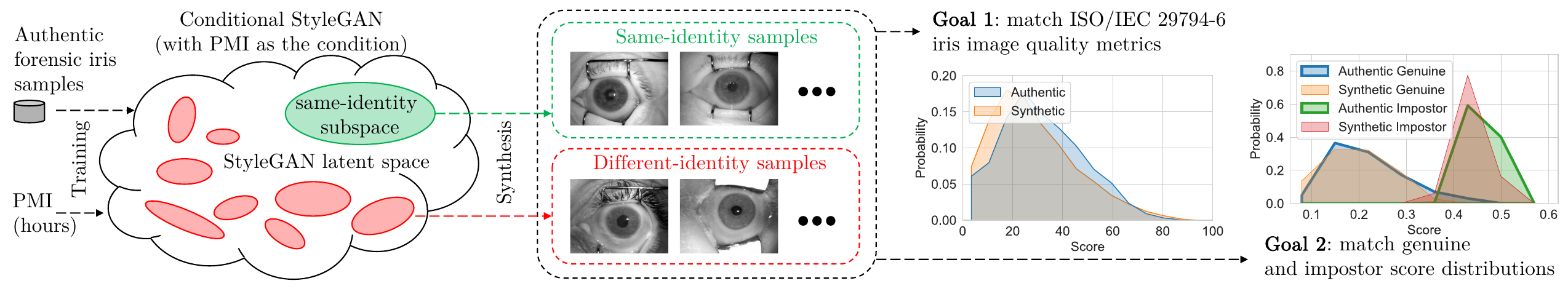}
    \caption{Synthesis of {\bf PMI-conditioned} same- and different-identity forensic iris images, with a {\bf desired iris image quality} (controlled by matching the ISO quality metrics), and with the {\bf desired genuine/impostor score distributions} (mimicking score distributions obtained for authentic forensic samples).}
    \label{fig:teaser}
\end{figure*}

%% file: sections/introduction.tex
The immeasurable appetite of modern machine learning for data is unfading, despite a success of various few-shot and transfer learning approaches. This is especially true in emerging domains, in which we are just discovering the domain's specificity, often solely through sparse and insufficient data. Post-mortem iris recognition is one of such domains. It gains more attention, however, after discovery that it may be considered in forensic applications \cite{trokielewicz2018iris}, and after seeing the first post-mortem-specific and human-interpretable recognition methods being proposed \cite{kuehlkamp2022interpretable,boyd2023human}. Although it sounds gruesome, it also constitutes yet another presentation attack type, and receives an attention in the presentation attack detection area.

The question arises how to progress this research knowing that collections of new post-mortem iris images will always be exceptionally complex and not creating well-balanced datasets in terms of the PMI (Post-Mortem Interval), environmental factors and/or subjects' demographical characteristics. We hypothesise that modern generative models, which proved useful in generating realistic biometric samples on non-existing identities, may partially fill this gap. However, an accurate modeling of post-mortem changes to the eye is virtually impossible due to multiple environmental, demise-reason-related, and subject's health factors that contribute to post-demise deformations of iris texture. This work proposes the first forensic iris synthesis model that generates samples matching selected statistics of authentic post-mortem samples, such as comparison score distributions and iris image quality metrics, as illustrated in Fig. \ref{fig:teaser}. More specifically, the {\bf main contribution of this work} is the conditional StyleGAN-based post-mortem iris synthesis model, which:

\begin{itemize}
    \item generates full-resolution ($640\times480$ pixels) iris post-mortem samples for a given post-mortem interval (up to 1674 hours),
    \item in addition to generation of new-identity images, also -- for a given non-existing identity -- generates an infinite number of samples considered as same-identity images; this was achieved by sampling from a vicinity of the conditional StyleGAN's latent vector (representing a given new identity) in order to mimic authentic genuine score distribution,
    \item generates post-mortem iris images with ISO quality metrics \cite{iso2014} mimicking those for authentic forensic iris images.
\end{itemize}

As an {\bf additional contribution}, facilitating the speed of adoption of this work, this paper offers a dataset of 180,000 synthetic post-mortem iris images, composed of 10,000 samples in each of the 18 PMI ranges (from 0 up to 1674 hours). For each PMI range there are 1,000 identities, and for each identity there are 10 same-identity (genuine) samples. We do not control the laterality (\ie left vs right iris image) when synthesizing samples. The source codes, model weights and instructions how to acquire a copy of the dataset are offered with this paper\footnote{\url{https://github.com/CVRL/Forensic-Iris-Image-Synthesis}}.

Two immediate applications of the results from this study include: (a) enhancement of the existing datasets for post-mortem iris recognition and iris presentation attack detection, and (b) supporting human forensic iris examiners' training, by generating variants of the same forensic sample to expose the experts to never-seen-before deformations of iris tissues.

%% file: sections/related-works.tex
The concept of synthesizing biometric iris images was, to our knowledge, first proposed by Cui \etal \cite{cui2004iris}, who leveraged principal component analysis and super-resolution techniques to create synthetic samples. Their approach set the stage for subsequent developments in iris image generation. Shah and Ross \cite{shah2006generating} harnessed Markov random fields to generate iris images. 
Zuo \etal \cite{zuo2007generation} introduced an anatomy-based model aimed at enhancing the realism of generated iris samples. 
Wei \etal \cite{wei2008synthesis} explored patch-based sampling as a means of generating iris images. Venugopalan and Savvides \cite{venugopalan2011generate} ventured into generating iris images from iris code templates, offering an alternative avenue for iris synthesis belonging to a family of ``inverse biometrics'' approaches. There have been several alternative methods proposed in the literature for the generation of iris images from binary templates \cite{galbally2013iris, galbally2016iris, wecker2005iris}. However, these methods often struggle to replicate the visual realism of generated iris images, resulting in synthetic samples that can appear artificial and thus can be easily detected by anomaly detection techniques.

In recent years, the Generative Adversarial Networks (GAN) have gained popularity for its success in realistic image generation, especially in tasks like face generation \cite{karras2017progressive, karras2019style, karras2020analyzing}. Some GAN-based methods have been explored for iris image generation. Minaee \etal \cite{minaee2018iris} utilized a Deep Convolutional GAN (DCGAN) to create realistic iris images, while Kohli \etal \cite{kohli2017synthetic} used DCGAN and iris quality metrics to study impact of synthetic irises on presentation attack detection. Yadav \etal \cite{yadav2019synthesizing} employed a relativistic average standard GAN (RaS-GAN) with Frechet Inception Distance to synthesize high-quality iris images. Yadav and Ross \cite{yadav2021cit} introduced a cyclic image translation GAN (CIT-GAN) for multi-domain style transfer in iris images. 

While the above GAN-based approaches have advanced the generation of realistic iris images, they still exhibit occasional distortions and blurry textures. Moreover, their primary use lies in anti-spoofing rather than iris recognition, as they cannot generate same-identity iris samples. To address this limitation, Kakani \etal \cite{kakani2023segmentation} proposed an iris synthesis model that is able to generate same-identity live iris samples. Maureira \etal \cite{maureira2021synthetic} utilized StyleGAN2 to generate synthetic periocular iris images. Wang \etal \cite{wang2022generating} harnessed the advanced StyleGAN2 architecture to create highly realistic iris images. To gain precise control over the generated attributes, they implemented contrastive learning, effectively disentangling identity vectors and enabling identity control for synthetic iris images.

To our knowledge, synthesis of both same- and different-identity {\it forensic} iris samples, for a given PMI, and matching both the comparison score and ISO image quality metrics to those of authentic post-mortem iris samples, have never been explored before. We hope that the model and approach offered in this paper will fill this gap.


%% file: sections/methodology.tex
\subsection{Dataset Composition and Splitting}
\label{sec:dataset}

There are currently three publicly available datasets of post-mortem iris images, and we have acquired all of them to conduct this research: 

\paragraph{Warsaw BioBase Post Mortem Iris v2.0 \cite{trokielewicz2018iris}} 
which contains 1,787 visible (VIS) and 1200 near-infrared (NIR) images from 37 deceased subjects. This dataset was collected in 2018 by conducting 1 to 13 sessions from 5 to 814 hours after death in the hospital mortuary. Most environmental conditions, such as where cadavers were kept before entering the cold storage, air pressure, and humidity, are unknown. However, the mortuary room temperature was approximately \ang{6} Celsius (\ang{42.8} Fahrenheit). 

\paragraph{Warsaw BioBase Post Mortem Iris v3.0 \cite{trokielewicz2020post}} which contains 785 VIS and 1094 NIR images from 42 cadavers collected 
in 2020 over a time horizon of up to 369 hours since demise. The environmental conditions were the same as in case of Warsaw BioBase Post Mortem Iris v2.0 set.

\paragraph{NIJ-2018-DU-BX-0215 \cite{Czajka2023software}} which is the latest and the largest forensic iris dataset, gathered at the Dutchess County Medical Examiner's Office (DCMEO) from 269 individuals who had passed away. A total of 53 data-gathering sessions were conducted and resulted in 
5,770 NIR images and 4,643 VIS images.

In this study, we only used NIR images and combined these three datasets into a cohesive set comprising 8,064 post-mortem iris images representing 338 subjects. 
The combined dataset is split into 18 classes, with each class gathering samples acquired within a 24-hour post-mortem interval (PMI). For example, Class 1 gathers samples acquired within the first 24 hours after death (so PMI $\in(0,24)$), Class 2 includes samples acquired from 25 to 48 hours after demise (so PMI $\in(25,48)$), and so on. This pattern continues up to Class 17, which includes samples with the PMI $\in(385,408)$. Due to limited availability of samples with PMI $>$ 408 hours, data for PMIs ranging from 409 to 1674 hours (max PMI for our combined dataset) were consolidated into Class 18, representing a more extended time since death. Table~\ref{tab:data-table} shows training data distribution across PMI classes, and Fig. \ref{fig:authentic-samples} shows example authentic samples from this combined dataset.

\begin{table}[!ht]
\centering
\footnotesize 
\caption{Number of training samples for each PMI range.}
\label{tab:data-table}
\begin{tabular}{lc|lc}
\toprule
Class \# & Number of & Class \# & Number of \\
(PMI range) & samples & (PMI range) & samples \\
\midrule
1 (0-24h) & 2490 & 10 (217h-240h) & 116 \\
2 (25h-48h) & 1542 & 11 (241h-264h) & 125 \\
3 (49h-72h) & 1094 & 12 (265h-288h) & 93 \\
4 (73h-96h) & 484 & 13 (289h-312h) & 45 \\
5 (97h-120h) & 494 & 14 (313h-336h) & 55 \\
6 (121h-144h) & 222 & 15 (337h-360h) & 64 \\
7 (145h-168h) & 328 & 16 (361h-384h) & 48 \\
8 (169h-192h) & 174 & 17 (385h-408h) & 54 \\
9 (193h-216h) & 238 & 18 (409h-1674h) & 327 \\
\bottomrule
\end{tabular}
\end{table}

\subsection{Synthetic Image Generator}
\label{sec:generator} 

Our approach is built on the StyleGAN2-ADA \cite{karras2020training} architecture, a model tailored for training generative adversarial networks with limited data. 
StyleGAN2-ADA employs non-leaking augmentations applied solely to the discriminator to diversify the data it sees, while the generator remains exposed only to the original samples, focusing on the genuine data distribution. Additionally, an adaptive mechanism dynamically controls the augmentation intensity, balancing between overfitting prevention and sample quality preservation. This strategy enables StyleGAN2-ADA to generate high-quality samples, even with limited training data effectively. The consideration of conditional information is made possible in the official implementation of StyleGAN2-ADA\footnote{\url{https://github.com/NVlabs/stylegan2-ada}}, which is used by us to generate forensic synthesis iris images with specific PMI range, given as the condition during synthesis.

The synthesis process initiates from a given latent code $\mathbf{z}$ in the latent space $\mathcal{Z}$. The mapping network in the original StyleGAN2-ADA generator \cite{karras2020analyzing}, represented as $f: \mathcal{Z} \to \mathcal{W}$, performs a projection of $\mathbf{z}$ into the intermediate latent code $\mathbf{w}$, where $\mathbf{w} \in \mathcal{W}$. In the conditional training scenario, $f$ possesses the capability to project $\mathbf{z}$ into $\mathbf{w}$ taking into account an embedded condition vector $\mathbf{y}$, 
which allows 
the generator to produce images with specific attributes 
based on the given condition. In our context, the condition vector represents the PMI range, which categorizes the time elapsed since death into bins, as described in Sec. \ref{sec:dataset}. This allows for learning the nature of iris texture deformations associated with different PMIs. The system thus enables creating synthetic iris images that statistically mimic the changes in iris appearance that occur over time after death.

Due to very sparse data available for training, a number of augmentations have been applied, including geometric and color adjustments, filtering, and image corruption techniques. Geometric transformations involve x-flips, translations, isotropic and anisotropic scaling, and arbitrary rotations. For intensity transformations, the pipeline adjusts brightness and contrast. 
Image-space filtering splits the frequency content into four bands and amplifies or weakens their amplitudes.

We trained the model in the original setting from scratch for a batch size of 32 and with authentic iris images rescaled to $256\times256$. During the generation of synthetic images, to synthesize same-identity iris images, we introduce additional noise to a randomly selected latent vector $\mathbf{w}$. Namely, for a given $\mathbf{w}$ we generate $\mathbf{w}' = \mathbf{\varepsilon w}$, where $\varepsilon\in(0,\varepsilon_{\max})$ and $\varepsilon_{\max}$ is the radius of a hypersphere in the latent space $\mathbf{w}$, found experimentally for a given dataset of post-mortem iris images, guarantying that resulting samples -- when matched to each other -- follow the same genuine score distribution as for authentic post-mortem samples.
In each generation of same-identity sample, $\varepsilon$ is drawn from a uniform distribution in a range of $(0,\varepsilon_{\max})$. Fig. \ref{fig:entire-genuine-auth-synth-score-dist} shows the genuine score distributions calculated for samples synthesized when drawing from hyperspheres in $\mathbf{w}$ of various radii $\varepsilon_{\max}$.

\input{figure-tex/entire-genuine-auth-synth}

\subsection{ISO/IEC 29794-6 Iris Image Quality Metrics}

In this study, we have implemented and calculated twelve iris image quality metrics defined in ISO/IEC FDIS 29794-6:2014(E) \cite{iso2014} to 
assess the usefulness of synthesized iris images for iris recognition and presentation attack detection. The selected metrics are:\\
    \null\hskip3mm\verb+USABLE_IRIS_AREA+,\\
    \null\hskip3mm\verb+IRIS_SCLERA_CONTRAST+, \\
    \null\hskip3mm\verb+IRIS_PUPIL_CONTRAST+, \\
    \null\hskip3mm\verb+PUPIL_BOUNDARY_CIRCULARITY+, \\
    \null\hskip3mm\verb+GREY_SCALE_UTILIZATION+, \\
    \null\hskip3mm\verb+IRIS_RADIUS+, \\
    \null\hskip3mm\verb+PUPIL_IRIS_RATIO+, \\
    \null\hskip3mm\verb+IRIS_PUPIL_CONCENTRICITY+, \\
    \null\hskip3mm\verb+MARGIN_ADEQUACY+, \\
    \null\hskip3mm\verb+SHARPNESS+, \\
    \null\hskip3mm\verb+MOTION_BLUR+, and \\
    \null\hskip3mm\verb+OVERALL_QUALITY+.

\vskip1mm
Except for \verb+GREY_SCALE_UTILIZATION+ (measured in bits), \verb+IRIS_RADIUS+ (measured in pixels), and \verb+MOTION_BLUR+ (measured as the major-to-minor axis ratio of the ellipse approximating the point spread function estimated from the image), all other metrics are in the range of (0,100). The 255 value means that a given image quality metric could not be calculated. 

\subsection{Iris Recognition Method}
\label{sec:matcher}
To calculate the comparison scores, we employed an open-source academic solution specially designed for post-mortem iris recognition, and based on human-driven binarized image features (HDBIF) \cite{czajka2019domain}. HDBIF combines deep learning-based iris segmentation and human-inspired domain-specific feature extraction to make iris biometrics more robust against post-mortem changes. This method uses domain-specific filters, derived via Independent Component Analysis (similarly to BSIF filters proposed by Kannala and Rahtu \cite{kannala2012bsif}), from iris image patches collected in an eye tracking experiment, in which subjects were solving a post-mortem iris recognition task. The resulting HDBIF filters are applied to normalized iris images and the convolutional results are binarized into a code, as in Daugman's approach \cite{daugman2007new}. Finally, the comparison score is calculated as the Hamming distance between codes representing non-occluded iris portions. The Hamming distance is simply the fraction of bits that the two iris codes disagree. This method is, to our knowledge, the state-of-the-art approach to post-mortem iris recognition \cite{boyd2023human}, hence its selection for this study

%% file: figure-tex/entire-genuine-auth-synth.tex
\begin{figure}[!ht]
    \centering
    \includegraphics[width=0.8\columnwidth]{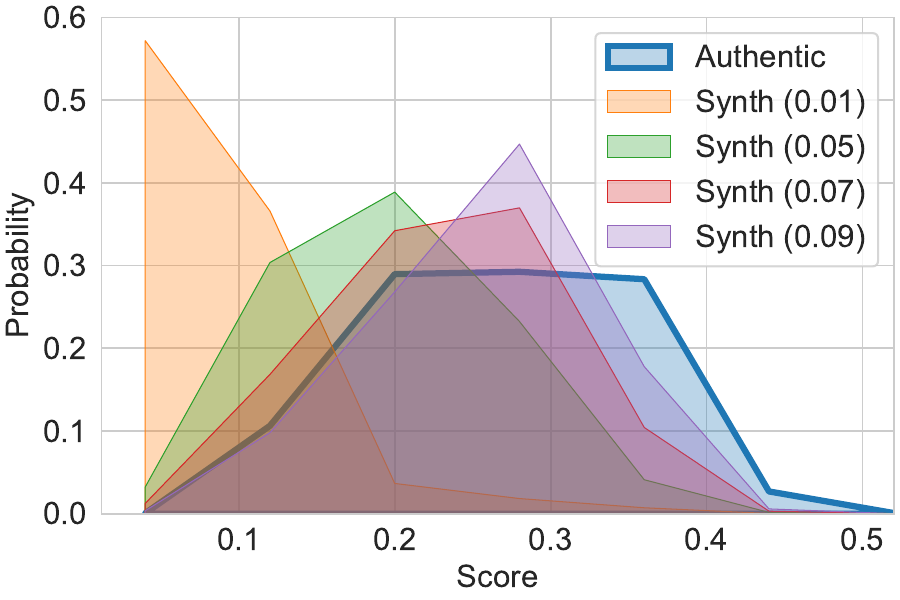}
    \caption{Genuine score distributions obtained for authentic post-mortem samples and synthesized post-mortem samples for different sizes of the hypersphere (defined by radius $\varepsilon_{\max}$, shown in parentheses) defining the same-identity manifold in the StyleGAN's latent space $\mathbf{w}$. The plots suggest that $\varepsilon_{\max}=0.09$ allows for mimicking the authentic genuine score distribution on average for all PMIs.}
    \label{fig:entire-genuine-auth-synth-score-dist}
\end{figure}

%% file: figure-tex/class-wise-authentic-synthetic-samples.tex
\begin{figure*}[!ht]
    \centering

    \begin{subfigure}{\linewidth}
        \includegraphics[width=\textwidth]{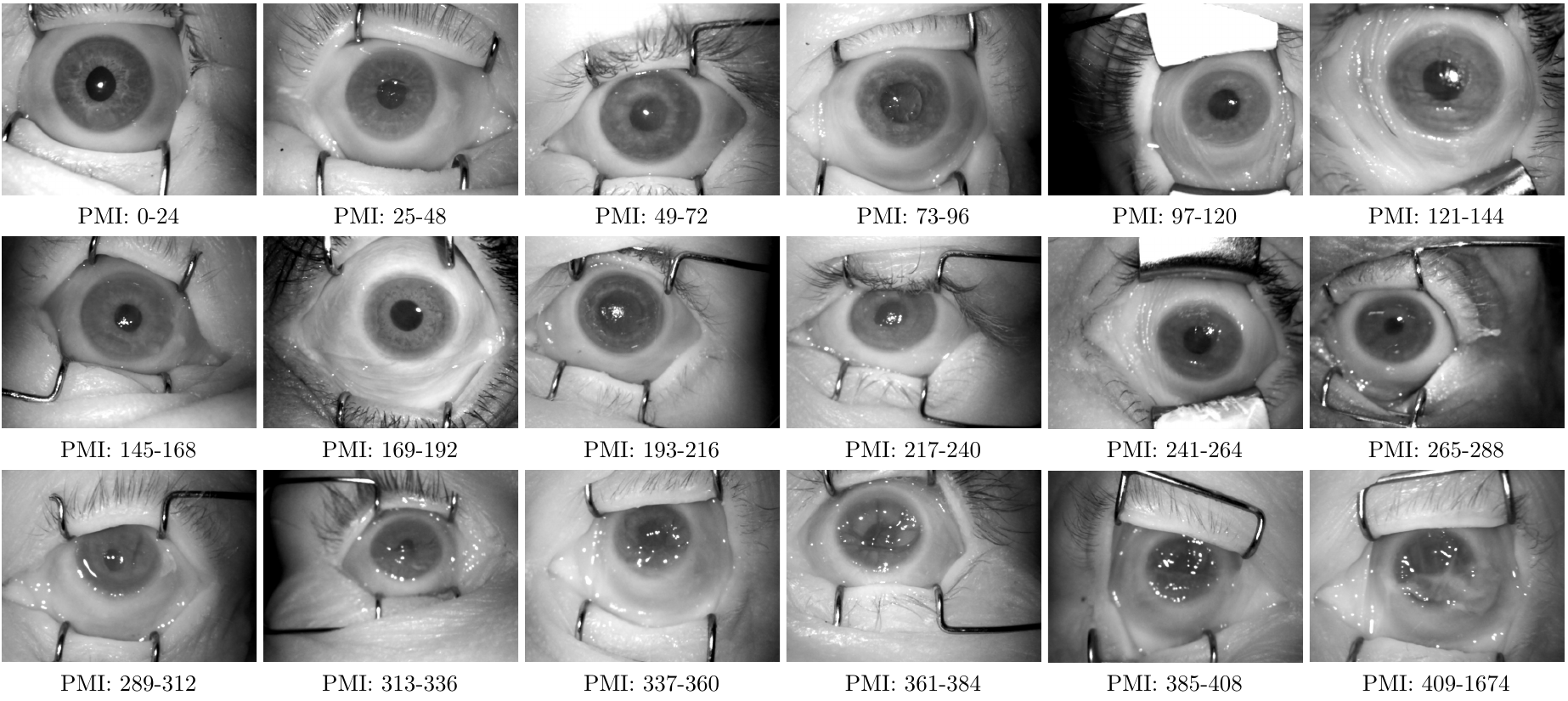}
        \caption{Example {\bf authentic} post-mortem iris images originating from the Warsaw BioBase Post Mortem Iris v2.0 \cite{trokielewicz2018iris} and v3.0 \cite{trokielewicz2020post} datasets (with approval).}
        \label{fig:authentic-samples}
    \end{subfigure}

    \begin{subfigure}{\linewidth}
        \includegraphics[width=\textwidth]{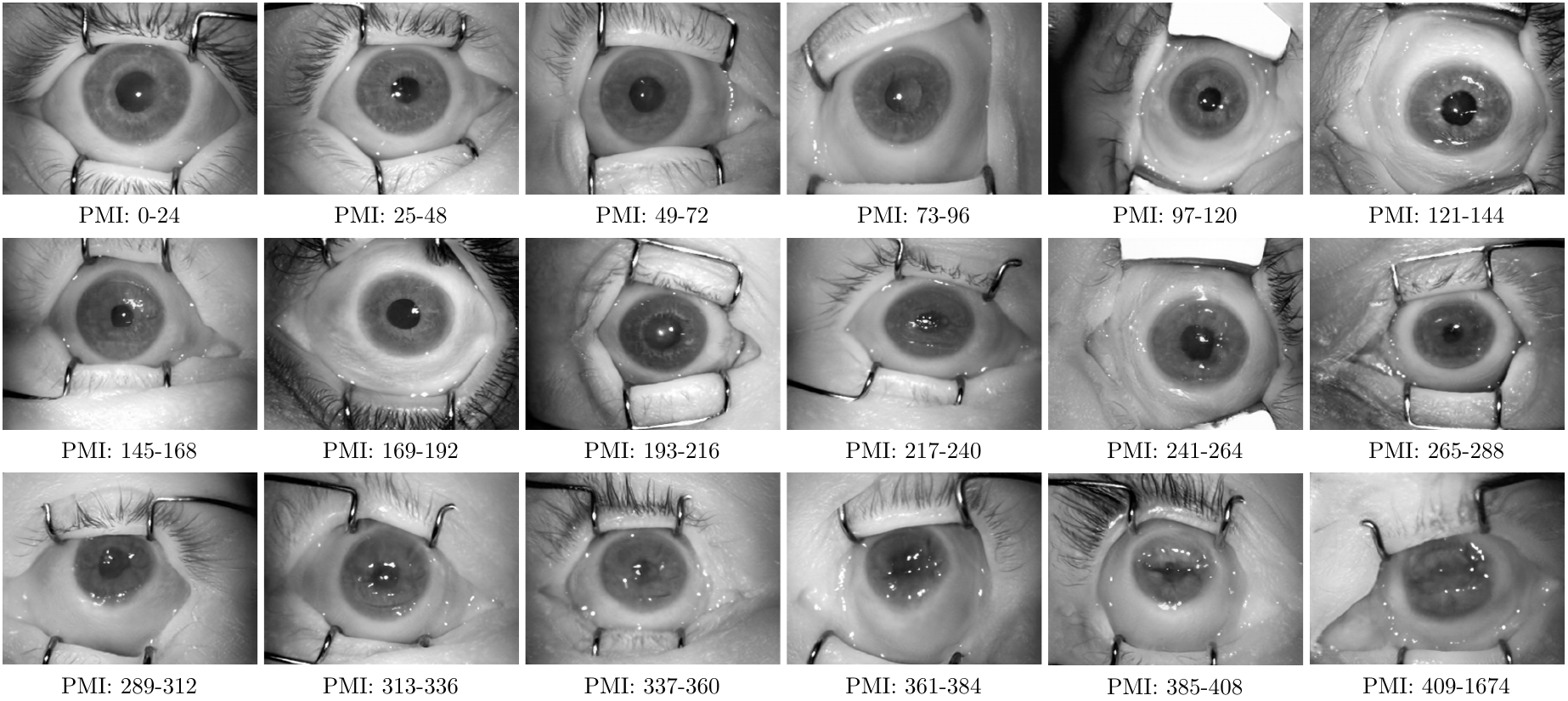}
        \caption{Example {\bf synthetic} samples, generated by the proposed model.}
        \label{fig:synthetic-samples}
    \end{subfigure}

    \caption{Examples of authentic and synthetically-generated images in the PMI ranges considered in this work. Note a remarkable realism of the generated synthetic images, and its capability to capture intricate details, such as the deformed (due to decomposition) iris texture, irregular specular highlights from the drying cornea (progressing when the PMI progresses), and even the metal eye retractors.}
    \label{fig:authentic-synthetic-samples}
\end{figure*}

%% file: figure-tex/noise-level-sample.tex
\begin{figure*}[!ht]
    \centering
    \includegraphics[width=\textwidth]{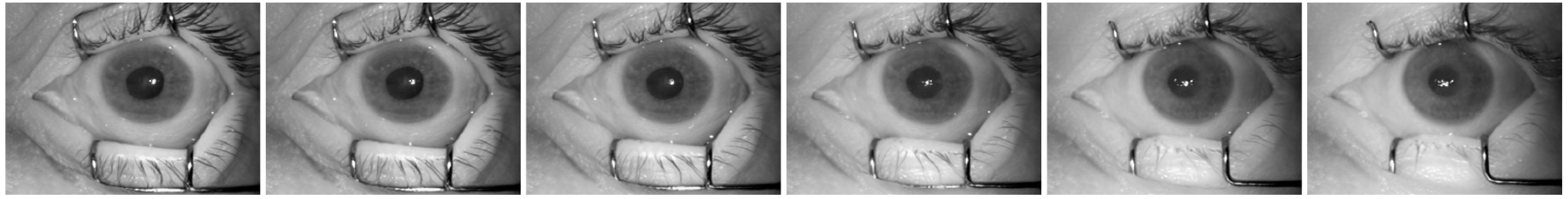}
    \caption{A set of synthetic images generated for a starting latent vector $\mathbf{w}$ (leftmost image) and latent vectors $\mathbf{w}' = \mathbf{\varepsilon w}$ sampled from the same-identity hyperspheres centered at $\mathbf{w}$ and of radii $\varepsilon = 0.01, 0.03, 0.05, 0.07, 0.09$ (from the second left to the rightmost image, respectively). 
    }
    \label{fig:noise-level-sample}
\end{figure*}

%% file: figure-tex/class-wise-score-dist.tex
\begin{figure*}[!ht] 
    \centering

    \begin{subfigure}{0.3\linewidth}
        \includegraphics[width=\linewidth]{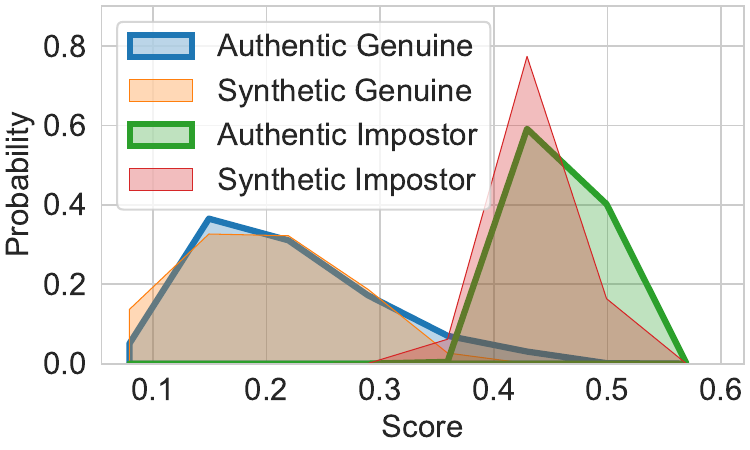}
        \caption{PMI: 0--24 hours}
        \label{fig:class1-score-dist}
    \end{subfigure}\hfill
    \begin{subfigure}{0.3\linewidth}
        \includegraphics[width=\linewidth]{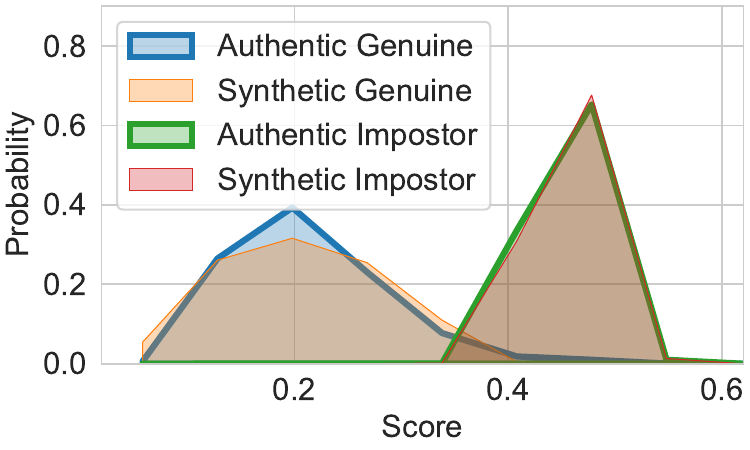}
        \caption{PMI: 25--48 hours}
        \label{fig:class2-score-dist}
    \end{subfigure}\hfill
    \begin{subfigure}{0.3\linewidth}
        \includegraphics[width=\linewidth]{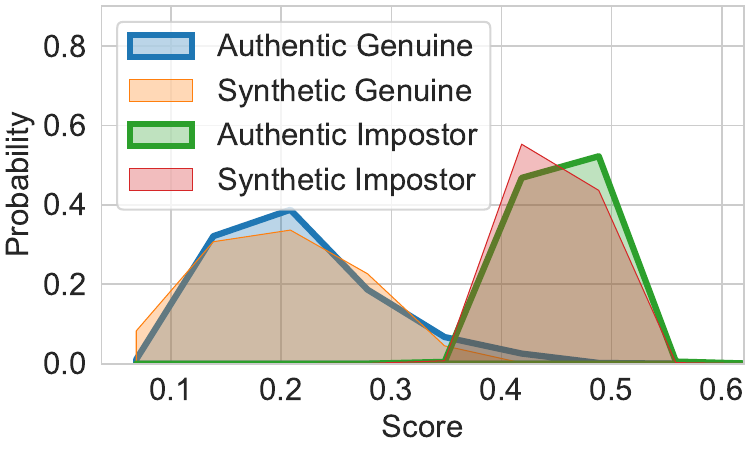}
        \caption{PMI: 49--72 hours}
        \label{fig:class3-score-dist}
    \end{subfigure}

    \begin{subfigure}{0.3\linewidth}
        \includegraphics[width=\linewidth]{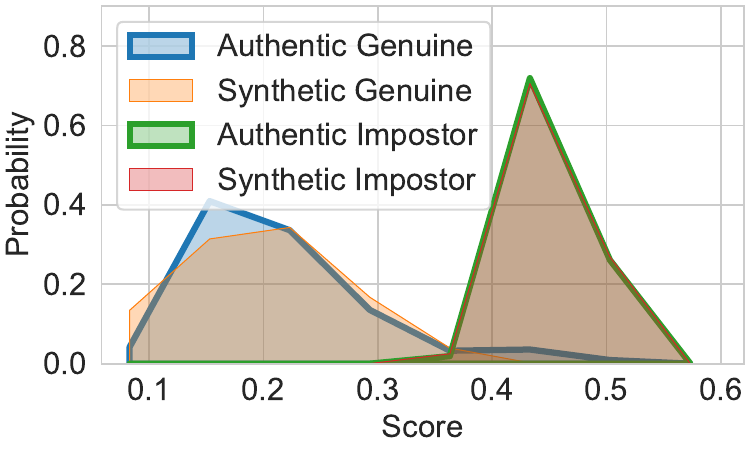}
        \caption{PMI: 73--96 hours}
        \label{fig:class4-score-dist}
    \end{subfigure}\hfill
    \begin{subfigure}{0.3\linewidth}
        \includegraphics[width=\linewidth]{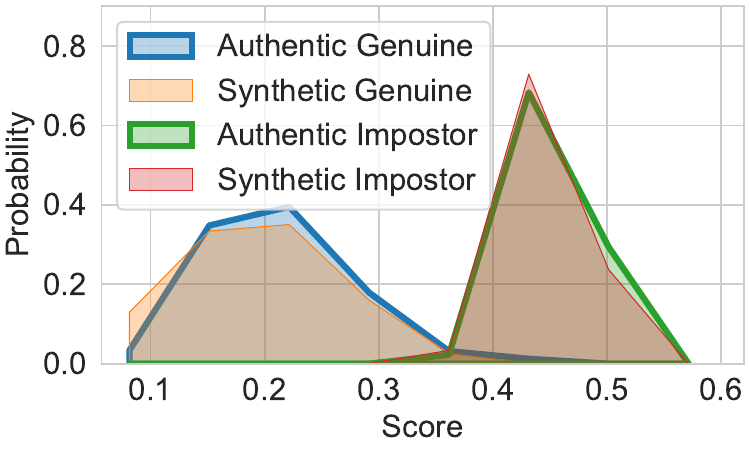}
        \caption{PMI: 97--120 hours}
        \label{fig:class5-score-dist}
    \end{subfigure}\hfill
    \begin{subfigure}{0.3\linewidth}
        \includegraphics[width=\linewidth]{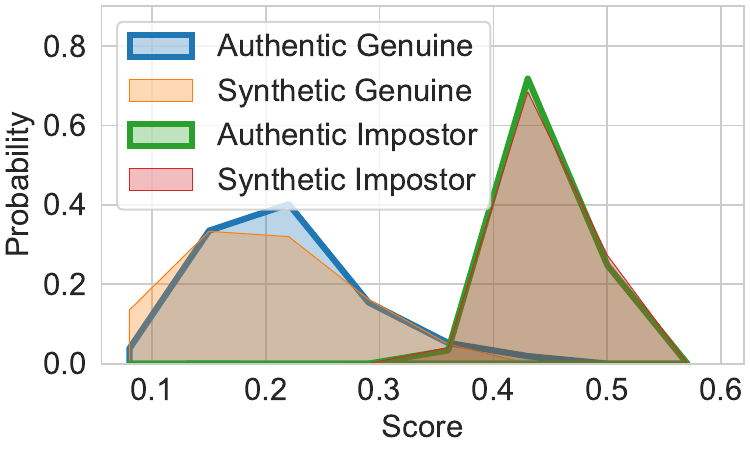}
        \caption{PMI: 121--144 hours}
        \label{fig:class6-score-dist}
    \end{subfigure}

    \begin{subfigure}{0.3\linewidth}
        \includegraphics[width=\linewidth]{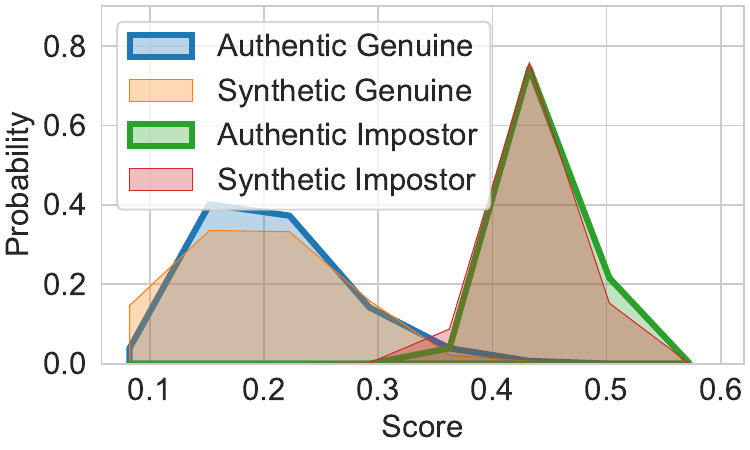}
        \caption{PMI: 145--168 hours}
        \label{fig:class7-score-dist}
    \end{subfigure}\hfill
    \begin{subfigure}{0.3\linewidth}
        \includegraphics[width=\linewidth]{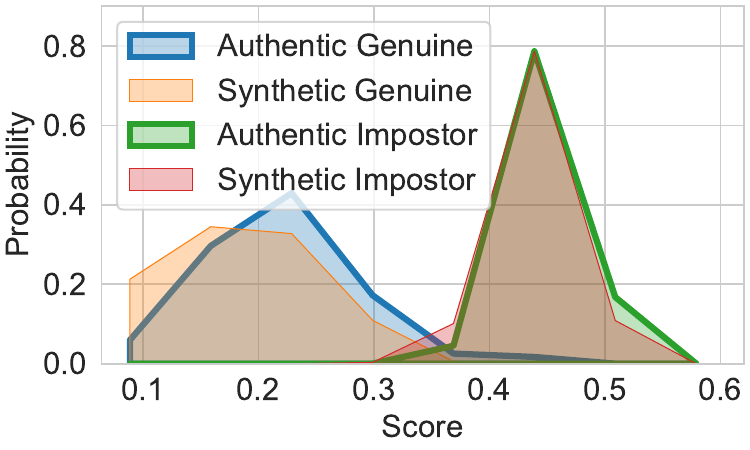}
        \caption{PMI: 169--192 hours}
        \label{fig:class8-score-dist}
    \end{subfigure}\hfill
    \begin{subfigure}{0.3\linewidth}
        \includegraphics[width=\linewidth]{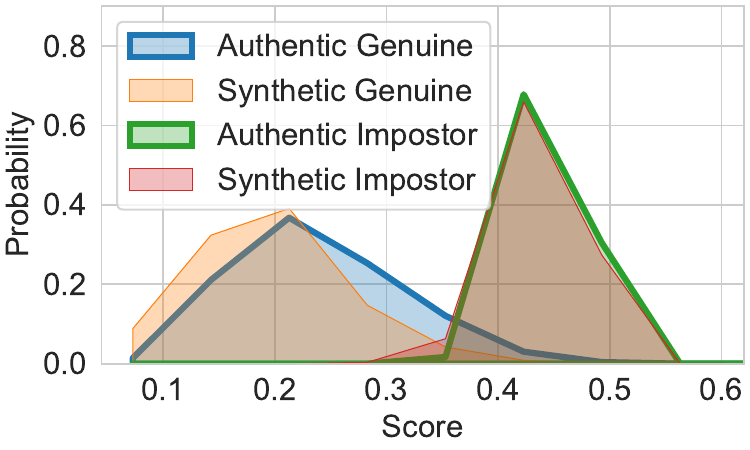}
        \caption{PMI: 193--216 hours}
        \label{fig:class9-score-dist}
    \end{subfigure}

    \begin{subfigure}{0.3\linewidth}
        \includegraphics[width=\linewidth]{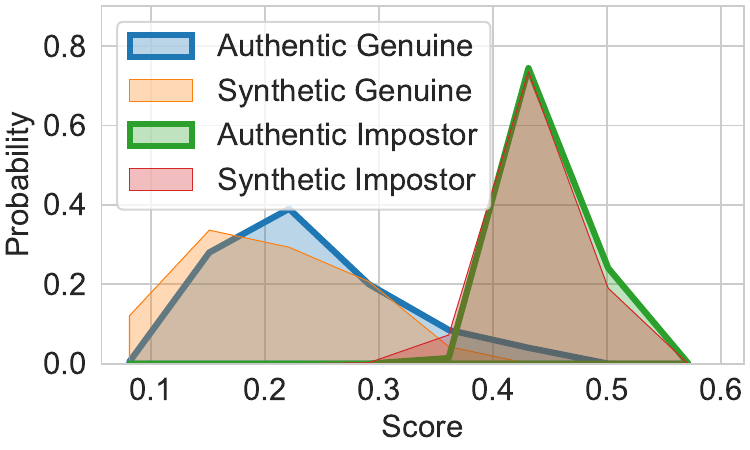}
        \caption{PMI: 217--240 hours}
        \label{fig:class10-score-dist}
    \end{subfigure}\hfill
    \begin{subfigure}{0.3\linewidth}
        \includegraphics[width=\linewidth]{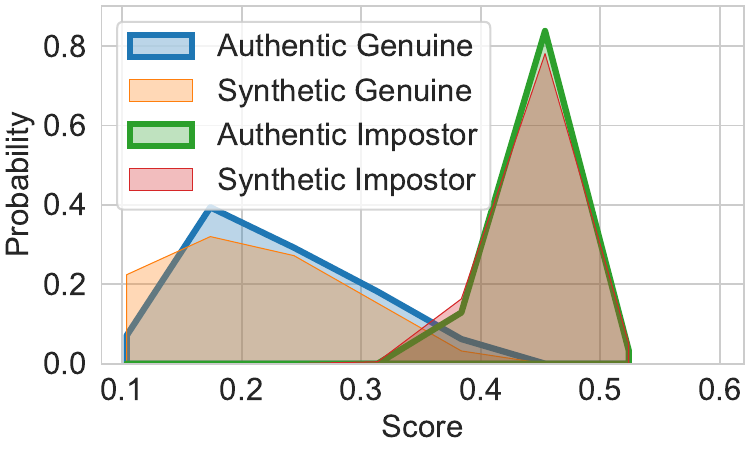}
        \caption{PMI: 241--264 hours}
        \label{fig:class11-score-dist}
    \end{subfigure}\hfill
    \begin{subfigure}{0.3\linewidth}
        \includegraphics[width=\linewidth]{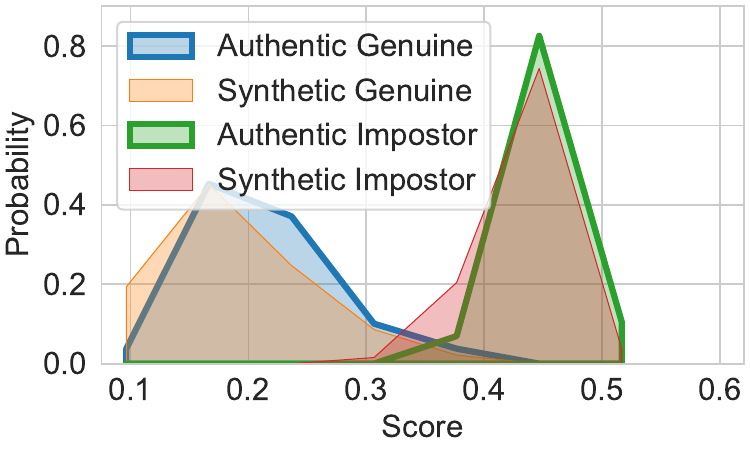}
        \caption{PMI: 265--288 hours}
        \label{fig:class12-score-dist}
    \end{subfigure}

    \begin{subfigure}{0.3\linewidth}
        \includegraphics[width=\linewidth]{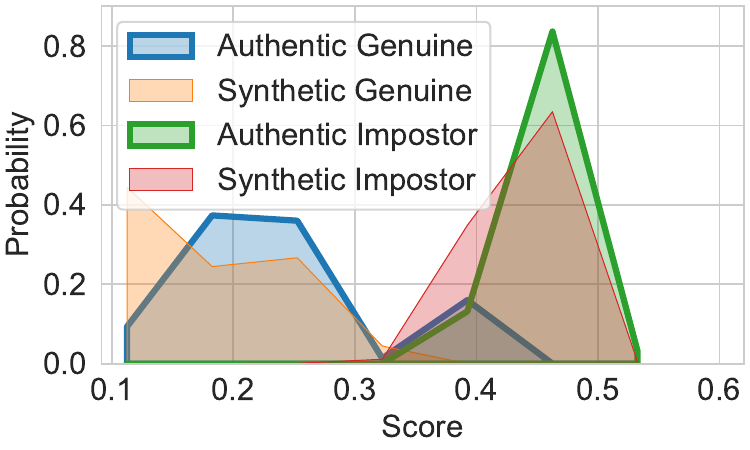}
        \caption{PMI: 289--312 hours}
        \label{fig:class13-score-dist}
    \end{subfigure}\hfill
    \begin{subfigure}{0.3\linewidth}
        \includegraphics[width=\linewidth]{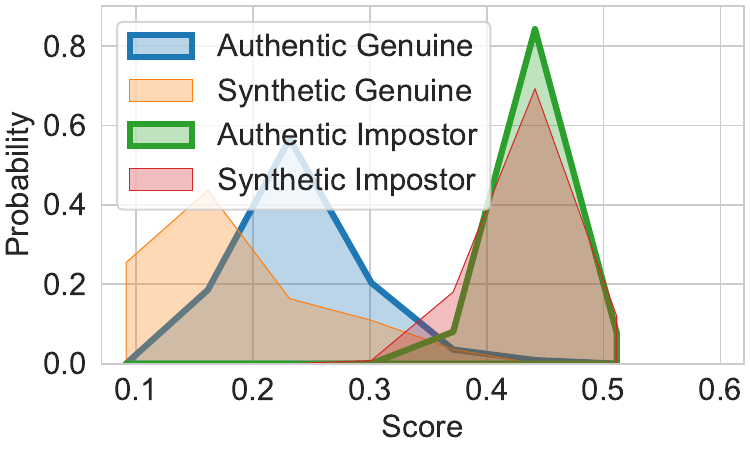}
        \caption{PMI: 313--336 hours}
        \label{fig:class14-score-dist}
    \end{subfigure}\hfill
    \begin{subfigure}{0.3\linewidth}
        \includegraphics[width=\linewidth]{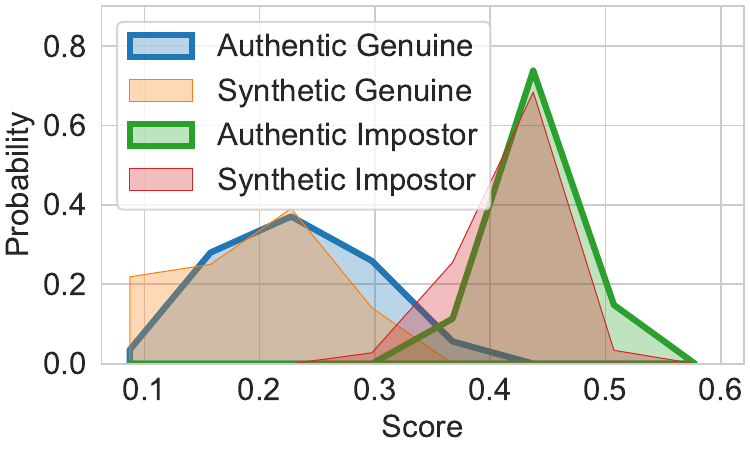}
        \caption{PMI: 337--360 hours}
        \label{fig:class15-score-dist}
    \end{subfigure}

    \begin{subfigure}{0.3\linewidth}
        \includegraphics[width=\linewidth]{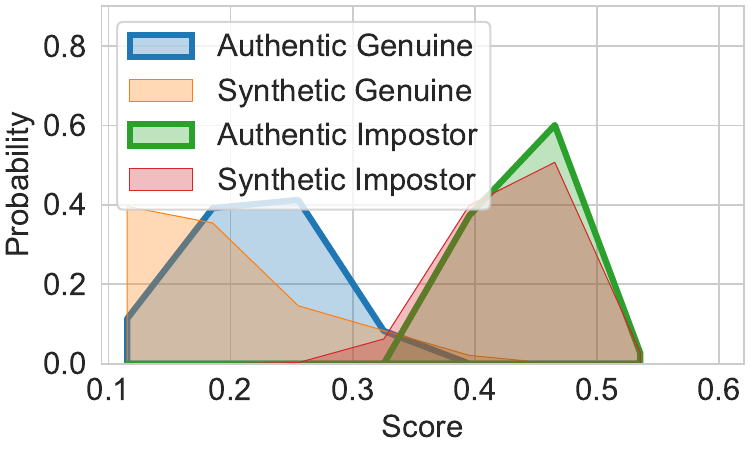}
        \caption{PMI: 361--384 hours}
        \label{fig:class16-score-dist}
    \end{subfigure}\hfill
    \begin{subfigure}{0.3\linewidth}
        \includegraphics[width=\linewidth]{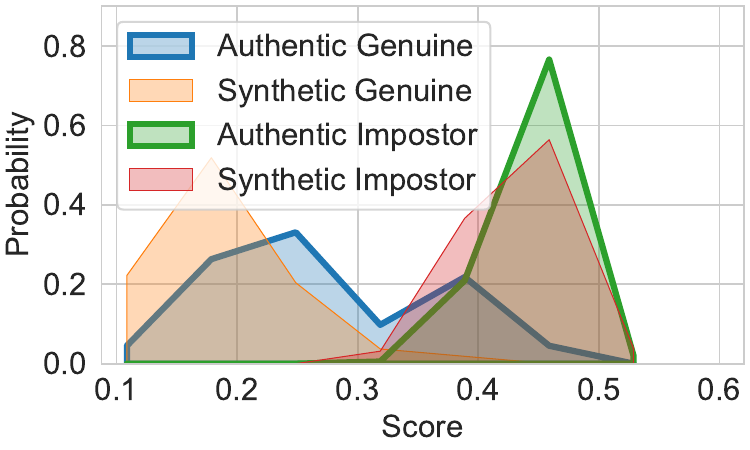}
        \caption{PMI: 385--408 hours}
        \label{fig:class17-score-dist}
    \end{subfigure}\hfill
    \begin{subfigure}{0.3\linewidth}
        \includegraphics[width=\linewidth]{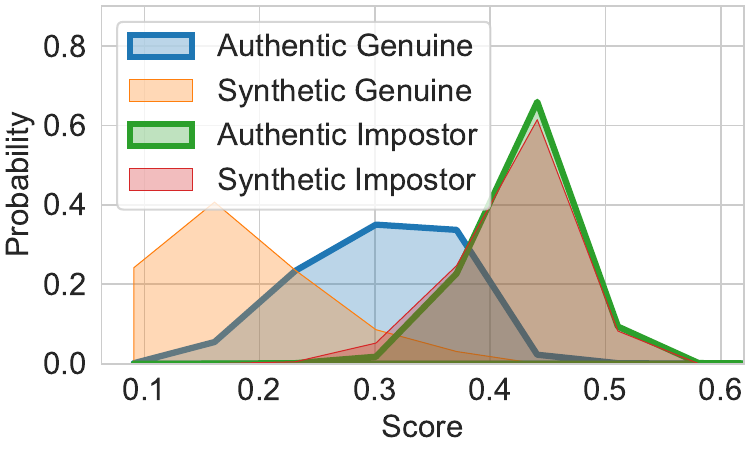}
        \caption{PMI: 409--1674 hours}
        \label{fig:class18-score-dist}
    \end{subfigure}

    \caption{Comparison score distributions obtained within all PMI ranges for authentic and synthetic post-mortem iris images.}
    \label{fig:class-wise-score-dist}
\end{figure*}

%% file: figure-tex/entire-data-iso-score-dist.tex
\begin{figure*}[!ht] 
    \centering

    \begin{subfigure}{0.33\linewidth}
        \includegraphics[width=\linewidth]{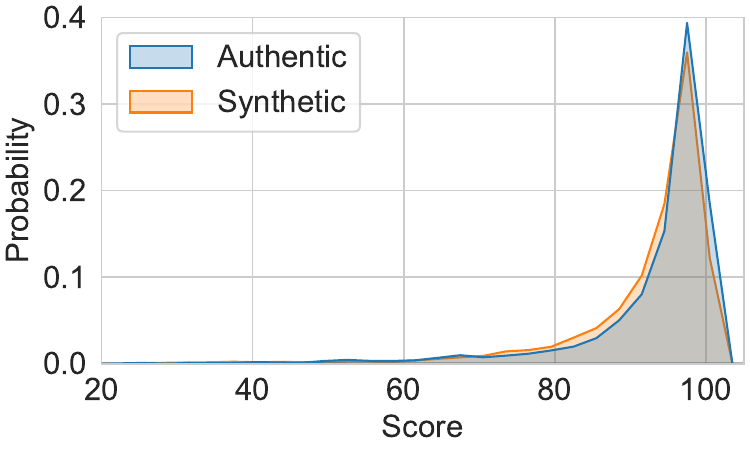}
        \caption{\texttt{USABLE\_IRIS\_AREA}}
        \label{fig:overall_USABLE_IRIS_AREA}
    \end{subfigure}\hfill
    \begin{subfigure}{0.33\linewidth}
        \includegraphics[width=\linewidth]{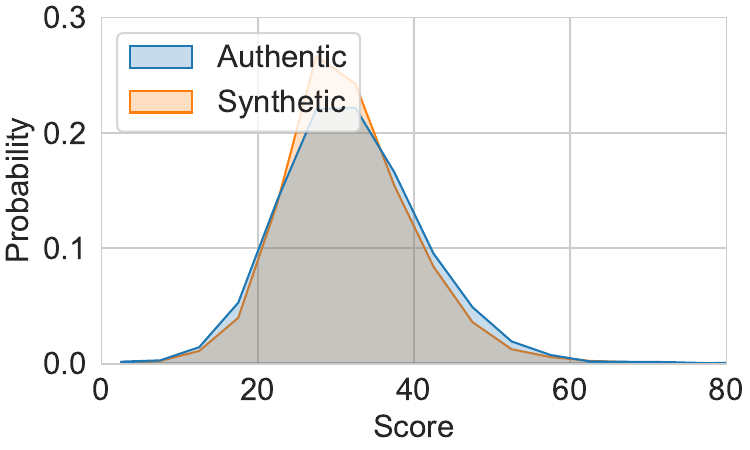}
        \caption{\texttt{IRIS\_SCLERA\_CONTRAST}}
        \label{fig:overall_IRIS_SCLERA_CONTRAST}
    \end{subfigure}\hfill
    \begin{subfigure}{0.33\linewidth}
        \includegraphics[width=\linewidth]{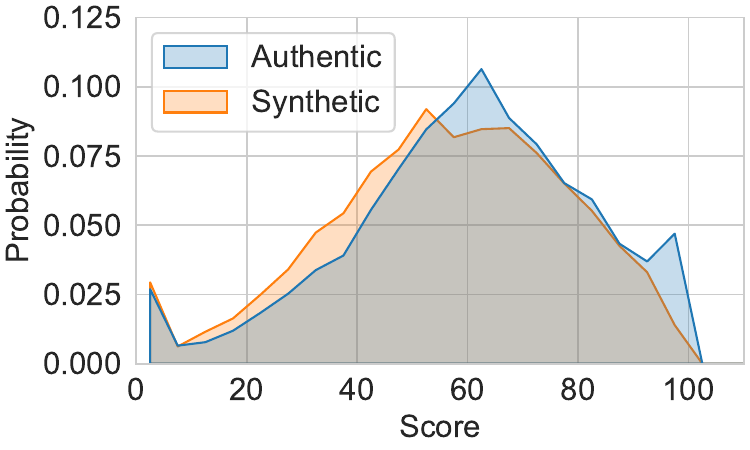}
        \caption{\texttt{IRIS\_PUPIL\_CONTRAST}}
        \label{fig:overall_IRIS_PUPIL_CONTRAST}
    \end{subfigure}
    
    \begin{subfigure}{0.33\linewidth}
        \includegraphics[width=\linewidth]{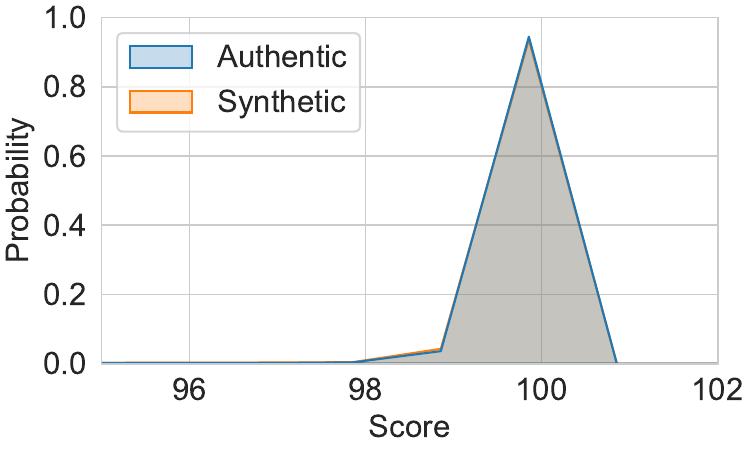}
        \caption{\texttt{PUPIL\_BOUNDARY\_CIRCULARITY}}
        \label{fig:overall_PUPIL_BOUNDARY_CIRCULARITY}
    \end{subfigure}\hfill
    \begin{subfigure}{0.33\linewidth}
        \includegraphics[width=\linewidth]{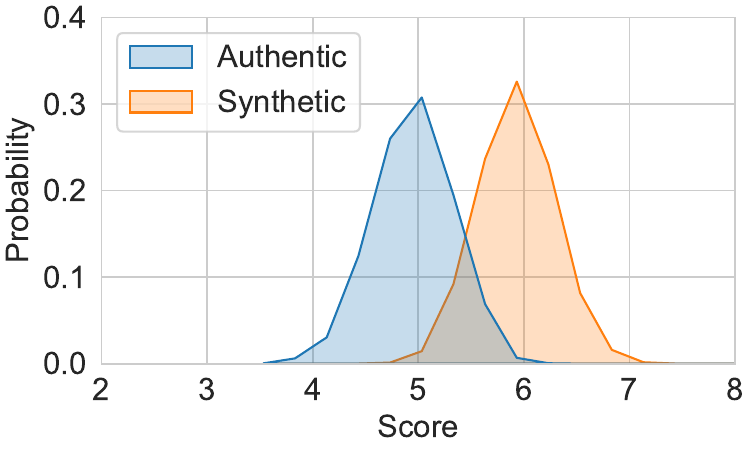}
        \caption{\texttt{GREY\_SCALE\_UTILISATION}}
        \label{fig:overall_GREY_SCALE_UTILISATION}
    \end{subfigure}\hfill
    \begin{subfigure}{0.33\linewidth}
        \includegraphics[width=\linewidth]{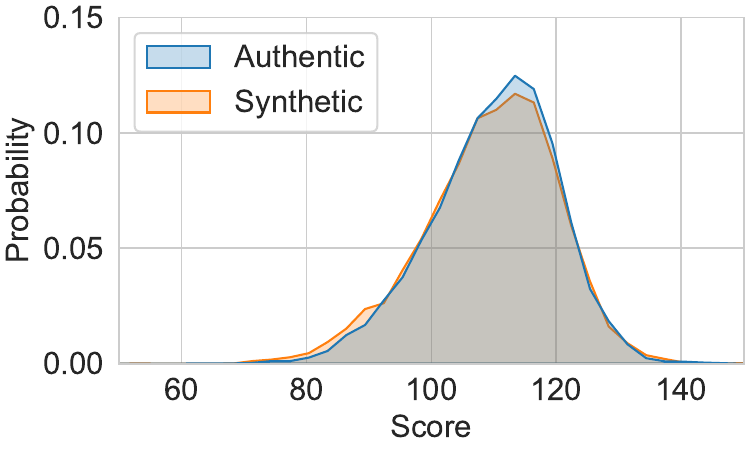}
        \caption{\texttt{IRIS\_RADIUS}}
        \label{fig:overall_IRIS_RADIUS}
    \end{subfigure}
    
    \begin{subfigure}{0.33\linewidth}
        \includegraphics[width=\linewidth]{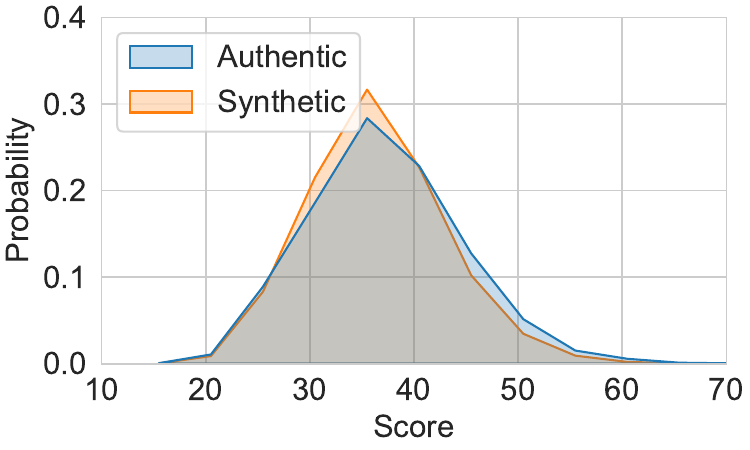}
        \caption{\texttt{PUPIL\_IRIS\_RATIO}}
        \label{fig:overall_PUPIL_IRIS_RATIO}
    \end{subfigure}\hfill
    \begin{subfigure}{0.33\linewidth}
        \includegraphics[width=\linewidth]{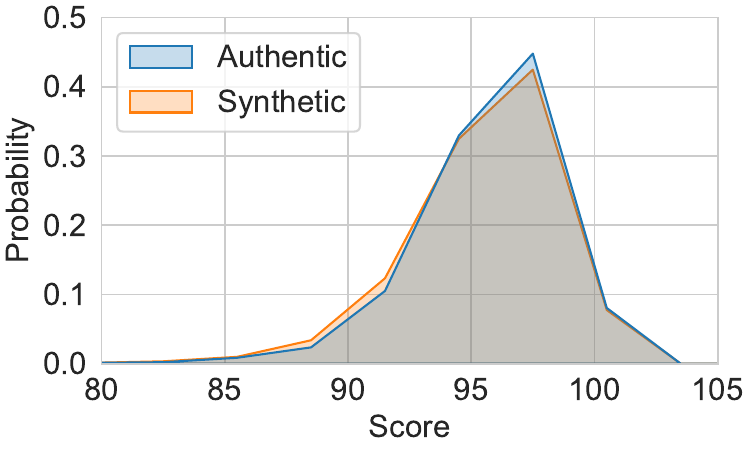}
        \caption{\texttt{IRIS\_PUPIL\_CONCENTRICITY}}
        \label{fig:overall_IRIS_PUPIL_CONCENTRICITY}
    \end{subfigure}\hfill
    \begin{subfigure}{0.33\linewidth}
        \includegraphics[width=\linewidth]{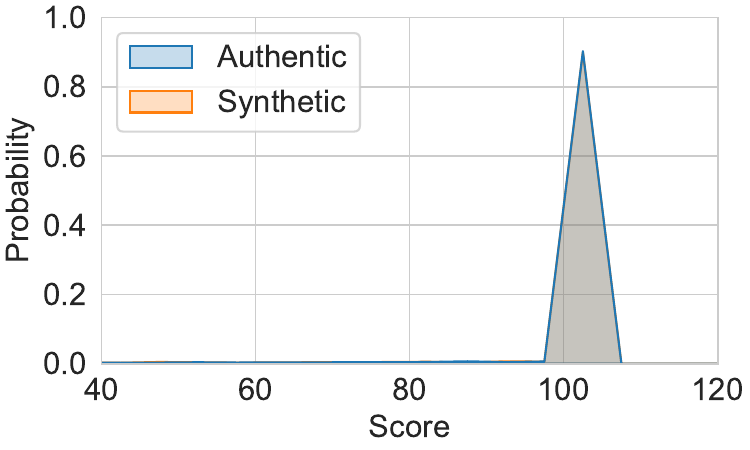}
        \caption{\texttt{MARGIN\_ADEQUACY}}
        \label{fig:overall_MARGIN_ADEQUACY}
    \end{subfigure}
    
    \begin{subfigure}{0.33\linewidth}
        \includegraphics[width=\linewidth]{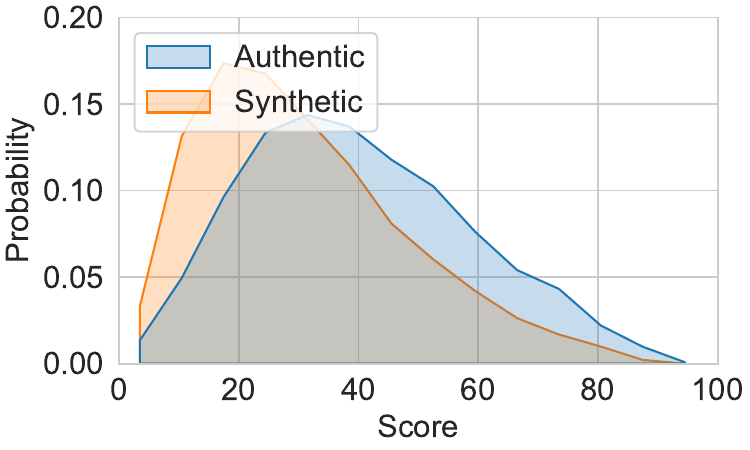}
        \caption{\texttt{SHARPNESS}}
        \label{fig:overall_SHARPNESS}
    \end{subfigure}\hfill
    \begin{subfigure}{0.33\linewidth}
        \includegraphics[width=\linewidth]{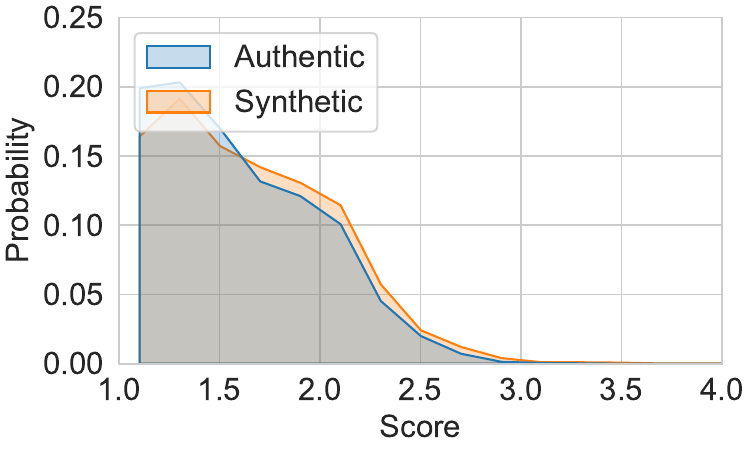}
        \caption{\texttt{MOTION\_BLUR}}
        \label{fig:overall_MOTION_BLUR}
    \end{subfigure}\hfill
    \begin{subfigure}{0.33\linewidth}
        \includegraphics[width=\linewidth]{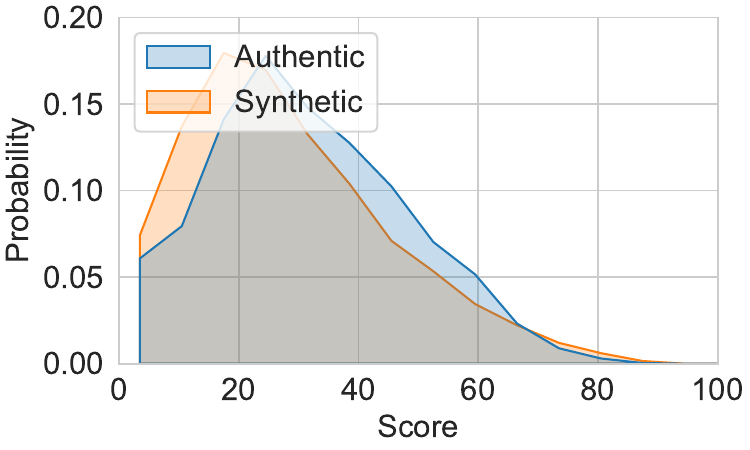}
        \caption{\texttt{OVERALL\_QUALITY}}
        \label{fig:overall_OVERALL_QUALITY}
    \end{subfigure}
    
    \caption{Distributions of the ISO/IEC 29794-6 iris image quality metrics calculated for authentic and synthesized post-mortem iris samples.}
    \label{fig:entire-data-iso-score-dist}
\end{figure*}

%% file: sections/experiment.tex
\subsection{Evaluation Samples} To evaluate the trained model in terms of its capabilities to generate ISO-compliant same- and different-identity post-mortem iris samples, we have generated synthetic forensic iris images representing all PMI ranges, and matching the number of samples in each PMI category with the numbers of authentic samples in the training set for each PMI class. 
Fig. \ref{fig:authentic-synthetic-samples} shows examples of authentic and synthetic images for each PMI class, where Fig. \ref{fig:authentic-samples} represents authentic samples and Fig. \ref{fig:synthetic-samples} features generated synthetic samples. 
A visual inspection suggests that samples generated by the proposed model include all intricate details specific to forensic iris images, such post-mortem-deformed iris tissue and cornea, metal retractors, eyelid shape due to application of such retractors, and deformed specular highlights caused by drying cornea.

\subsection{Matching the Comparison Score Distributions}
As described in Sec. \ref{sec:generator}, to generate synthetic images that closely mimic authentic images, we sampled from hyperspheres of different sizes, defined by a radius $\varepsilon_{\max}$ centered at a latent vector representing a given new identity. 
As we can see in Figure~\ref{fig:noise-level-sample}, depending on the distance from the starting latent vector, the synthesis process introduces variations in the iris texture, pupil size, and level of the decomposition-related deformations. The obvious question is thus about the radius $\varepsilon_{\max}$ to not go ``too far'' if one wants to generate genuine samples. 

To assess the $\varepsilon_{\max}$ value in a quantitative manner, we calculated the genuine comparison score distributions (using the HDBIF matcher, Sec. \ref{sec:matcher}) for both the authentic and synthetic samples to see which $\varepsilon_{\max}$ levels allow to align the authentic and synthetic genuine score distributions. An interesting observation was that, as shown in Fig. \ref{fig:entire-genuine-auth-synth-score-dist}, $\varepsilon_{\max}=0.09$ allows for a close alignment of these two distributions for the entire dataset. However, when we consider the distribution of genuine scores on a class-wise basis, 
setting a PMI-specific $\varepsilon_{\max}$ may be necessary. For instance $\varepsilon_{\max} = 0.05$ is a better choice for PMIs up to 312h, and $\varepsilon_{\max} = 0.09$ allows for a better score distribution alignment for the remaining PMI ranges (313h-336h, 337h-360h, 361h-384h, 384h-408h, and 409h-1674h). This phenomenon can be explained by large and unpredictable dynamics of post-mortem changes for eyes with larger PMI values. Fig.~\ref{fig:class-wise-noise-level-score-dist} in the supplementary materials illustrates all genuine score distributions calculated for authentic images and synthetic samples generated for various $\varepsilon_{\max}$ and for each PMI class.

For simplicity and illustration of the model's operation, a single $\varepsilon_{\max} = 0.05$ was used to generate samples for the final evaluation, illustrated for all PMI classes in Fig.~\ref{fig:class-wise-score-dist}.

\subsection{Matching the ISO/IEC 29794-6 Image Quality Metrics}

To assess the iris recognition-specific quality of the generated images, we analyzed how eleven ISO/IEC 29794-6 iris image quality metrics align between the authentic and synthetic images. These metrics offer a multifaceted evaluation, considering aspects such as iris pattern visibility, pupil-to-iris size ratio, pupil shape regularity, gray-scale utilization, and sharpness. In addition, the overall quality metric, combining those eleven properties, has been calculated. A visual representation of the distributions of ISO metrics allowing to assess the quality of the synthesized images in relation to the authentic samples is shown in Fig.~\ref{fig:entire-data-iso-score-dist}. 
The synthetic images closely mirrored the quality properties of the authentic images. This congruence in distributions highlights the model's proficiency in replicating various quality attributes. However, two discernible differences were found. The first one in the \verb+GREY_SCALE_UTILIZATION+ metric, where the synthetic images exhibited a higher score for most of the generated samples compared to authentic images, indicating a more efficient utilization of the grayscale spectrum by the trained model. These variations may be partially attributed to the extensive color transformations applied within the adaptive discriminator augmentation pipeline of the StyleGAN2-ADA model. The second difference can be noted in \verb+SHARPNESS+ values, being a bit smaller for synthetic samples. This can be attributed to up-scaling carried out after the synthesis is finished. Knowing that iris recognition usually employs lower-pass filters to encode identity features, this should not have a significant impact on the usability of the generated samples. 

%% file: sections/conclusion.tex
This study offers a conditional StyleGAN-based generative method to synthesize post-mortem iris images, with a hope to advance both forensic iris recognition (including human examination) and iris presentation attack detection, by offering the first-of-this-kind tool to generate never-seen-before, realistic, ISO-compliant, identity-preserving samples exhibiting tissue deformations typical for a given post-mortem interval. The same-identity samples were possible to be generated by an appropriate definition of latent space sub-manifolds ``storing'' representations of the same identities. These sub-manifolds were discovered by matching the genuine score distributions calculated for synthetic and authentic images. In addition to the sources codes and model weights, this paper also offers a ready-to-use dataset of 180,000 synthetic, post-mortem iris images for 18 different PMI ranges. 


%% file: figure-tex/class-wise-noise-level-score-dist.tex
\begin{figure*}[!ht]
    \centering
        \begin{subfigure}{0.3\linewidth}
            \includegraphics[width=\linewidth]{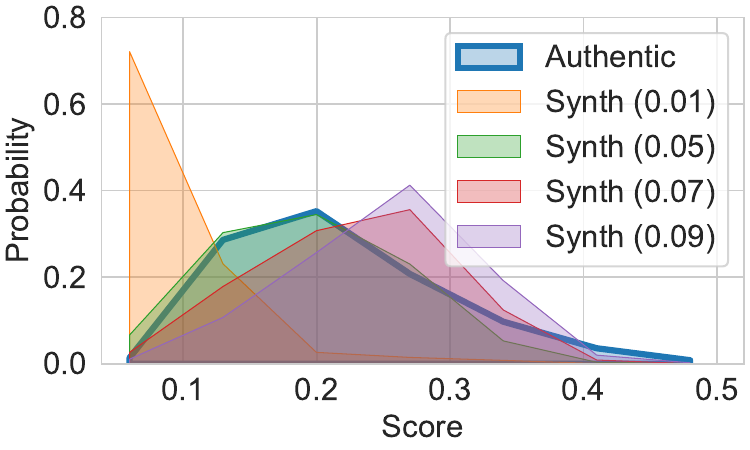}
            \caption{PMI: 0--24 hours}
            \label{fig:class1-noise-score-dist}
        \end{subfigure}\hfill
        \begin{subfigure}{0.3\linewidth}
            \includegraphics[width=\linewidth]{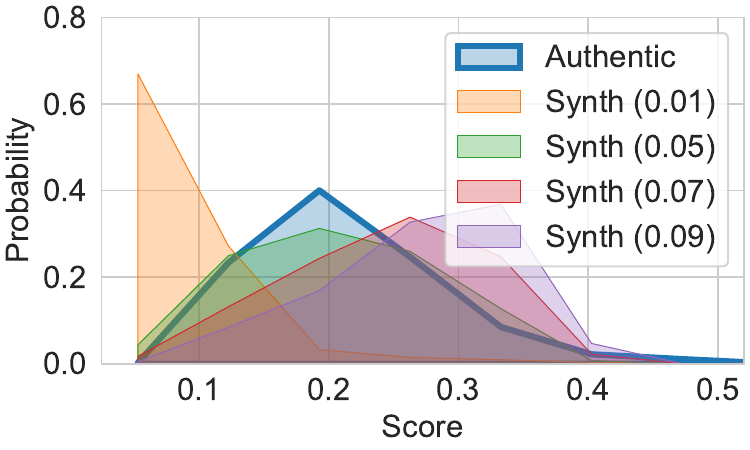}
            \caption{PMI: 25--48 hours}
            \label{fig:class2-noise-score-dist}
        \end{subfigure}\hfill
        \begin{subfigure}{0.3\linewidth}
            \includegraphics[width=\linewidth]{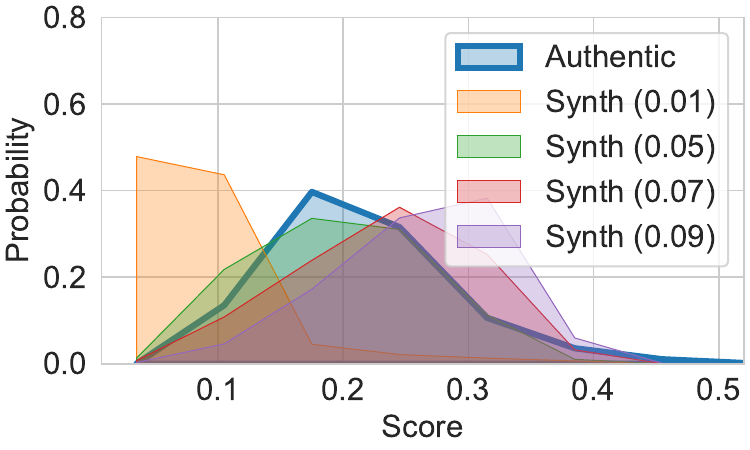}
            \caption{PMI: 49--72 hours}
            \label{fig:class3-noise-score-dist}
        \end{subfigure}\hfill
    
        \begin{subfigure}{0.3\linewidth}
            \includegraphics[width=\linewidth]{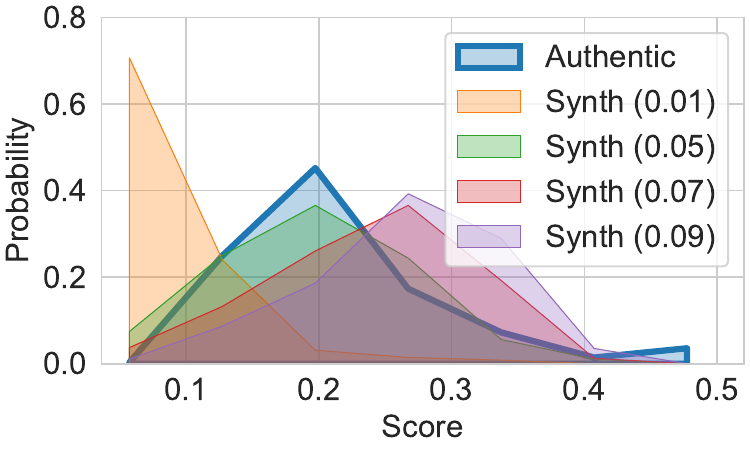}
            \caption{PMI: 73--96 hours}
            \label{fig:class4-noise-score-dist}
        \end{subfigure}\hfill
        \begin{subfigure}{0.3\linewidth}
            \includegraphics[width=\linewidth]{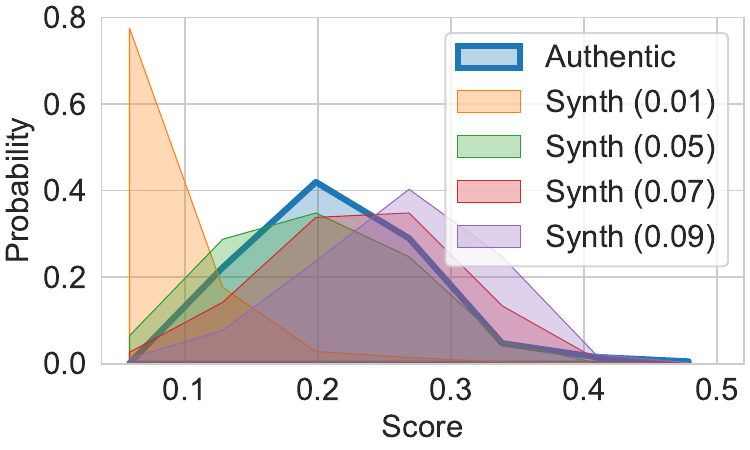}
            \caption{PMI: 97--120 hours}
            \label{fig:class5-noise-score-dist}
        \end{subfigure}\hfill
        \begin{subfigure}{0.3\linewidth}
            \includegraphics[width=\linewidth]{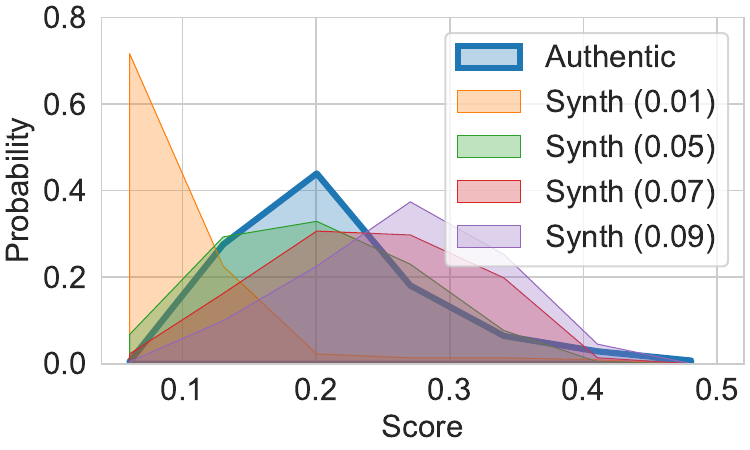}
            \caption{PMI: 121--144 hours}
            \label{fig:class6-noise-score-dist}
        \end{subfigure}
    
        \begin{subfigure}{0.3\linewidth}
            \includegraphics[width=\linewidth]{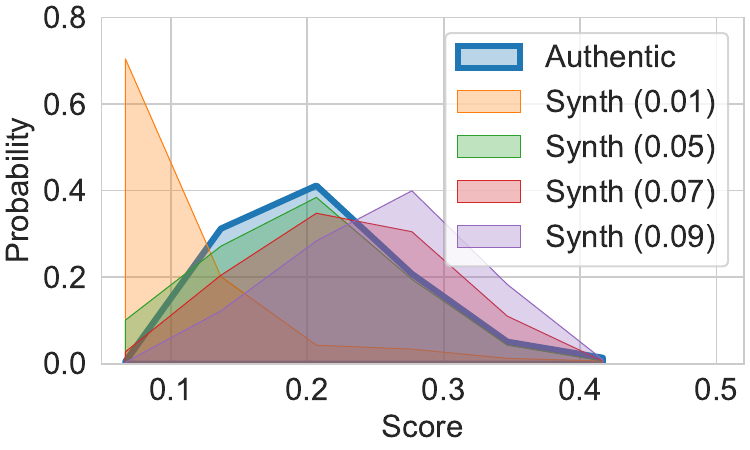}
            \caption{PMI: 145--168 hours}
            \label{fig:class7-noise-score-dist}
        \end{subfigure}\hfill
        \begin{subfigure}{0.3\linewidth}
            \includegraphics[width=\linewidth]{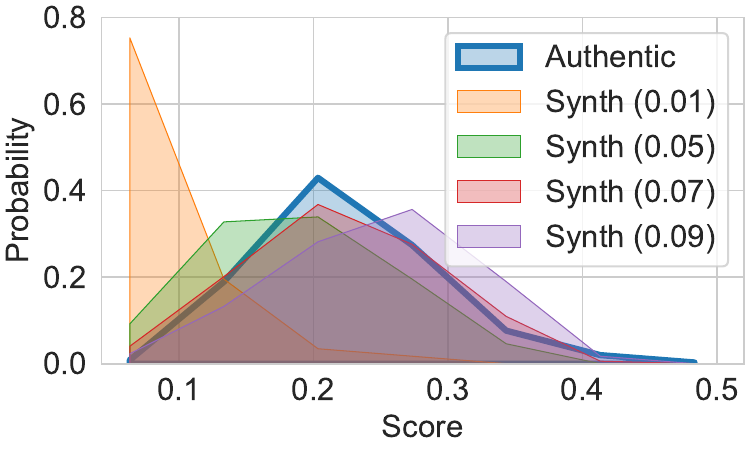}
            \caption{PMI: 169--192 hours}
            \label{fig:class8-noise-score-dist}
        \end{subfigure}\hfill
        \begin{subfigure}{0.3\linewidth}
            \includegraphics[width=\linewidth]{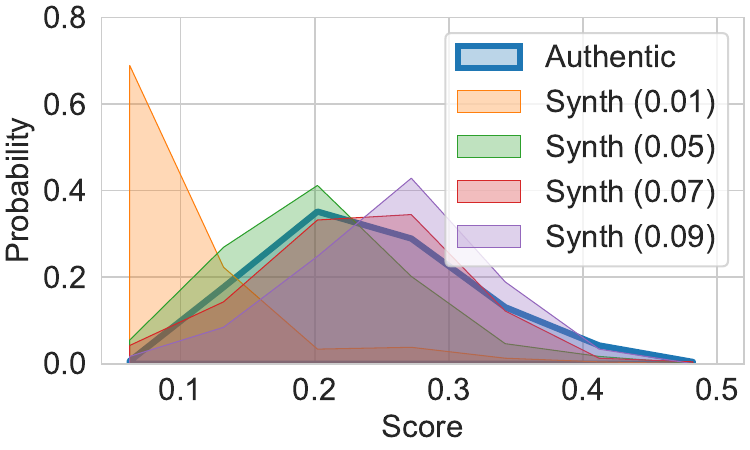}
            \caption{PMI: 193--216 hours}
            \label{fig:class9-noise-score-dist}
        \end{subfigure}

        \begin{subfigure}{0.3\linewidth}
            \includegraphics[width=\linewidth]{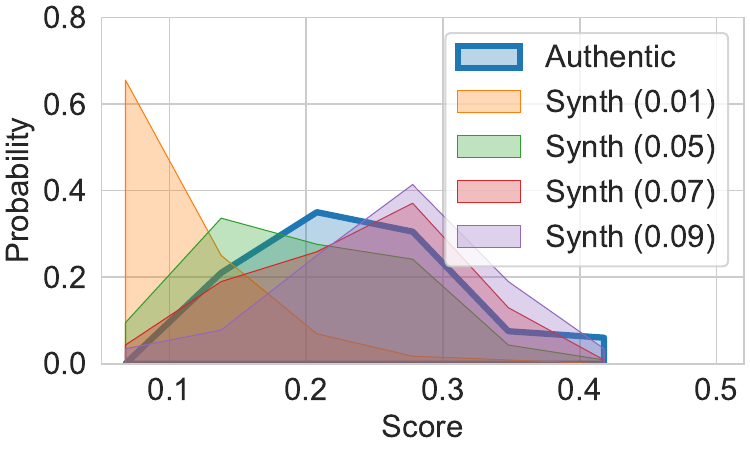}
            \caption{PMI: 217--240 hours}
            \label{fig:class10-noise-score-dist}
        \end{subfigure}\hfill
        \begin{subfigure}{0.3\linewidth}
            \includegraphics[width=\linewidth]{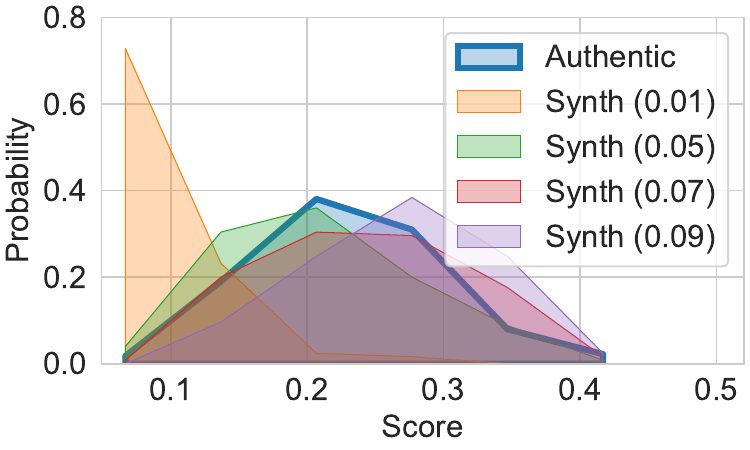}
            \caption{PMI: 241--264 hours}
            \label{fig:class11-noise-score-dist}
        \end{subfigure}\hfill
        \begin{subfigure}{0.3\linewidth}
            \includegraphics[width=\linewidth]{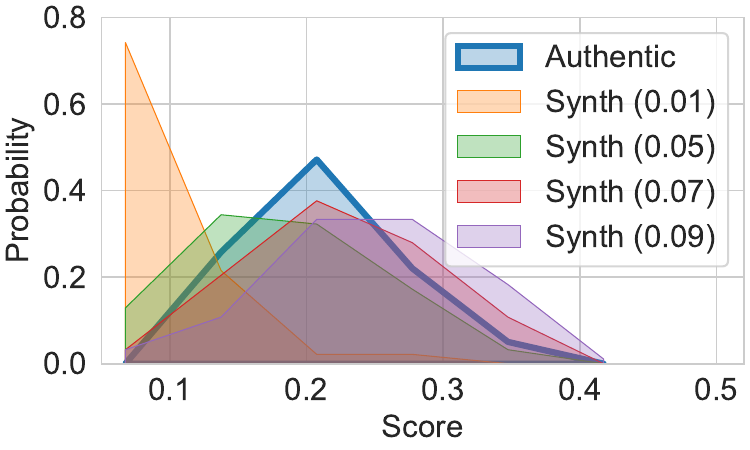}
            \caption{PMI: 265--288 hours}
            \label{fig:class12-noise-score-dist}
        \end{subfigure}
    
        \begin{subfigure}{0.3\linewidth}
            \includegraphics[width=\linewidth]{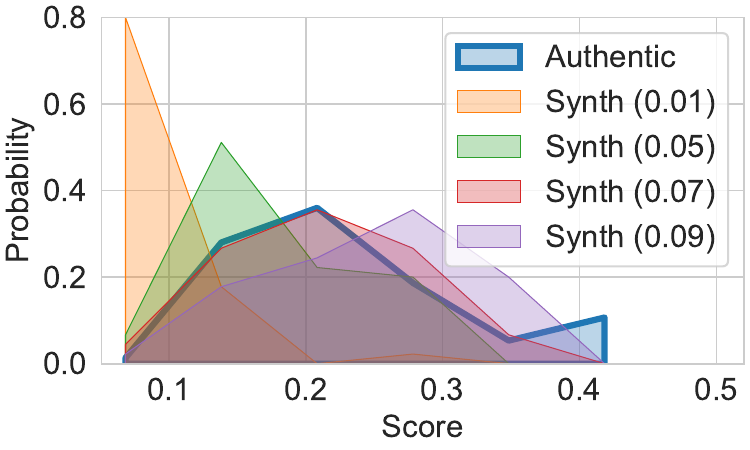}
            \caption{PMI: 289--312 hours}
            \label{fig:class13-noise-score-dist}
        \end{subfigure}\hfill
        \begin{subfigure}{0.3\linewidth}
            \includegraphics[width=\linewidth]{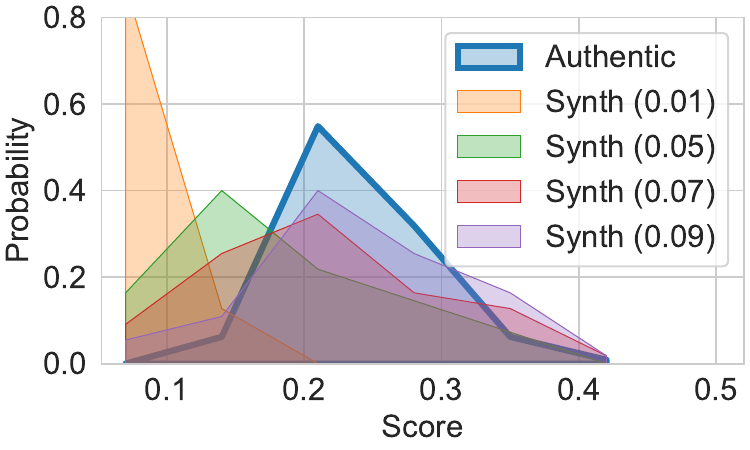}
            \caption{PMI: 313--336 hours}
            \label{fig:class14-noise-score-dist}
        \end{subfigure}\hfill
        \begin{subfigure}{0.3\linewidth}
            \includegraphics[width=\linewidth]{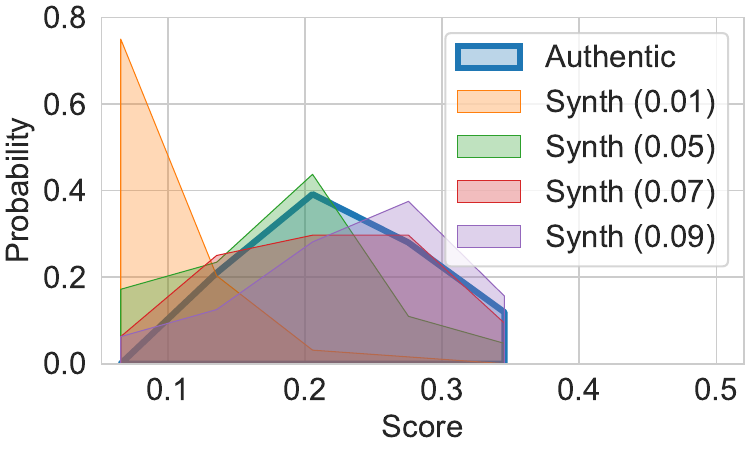}
            \caption{PMI: 337--360 hours}
            \label{fig:class15-noise-score-dist}
        \end{subfigure}

        \begin{subfigure}{0.3\linewidth}
            \includegraphics[width=\linewidth]{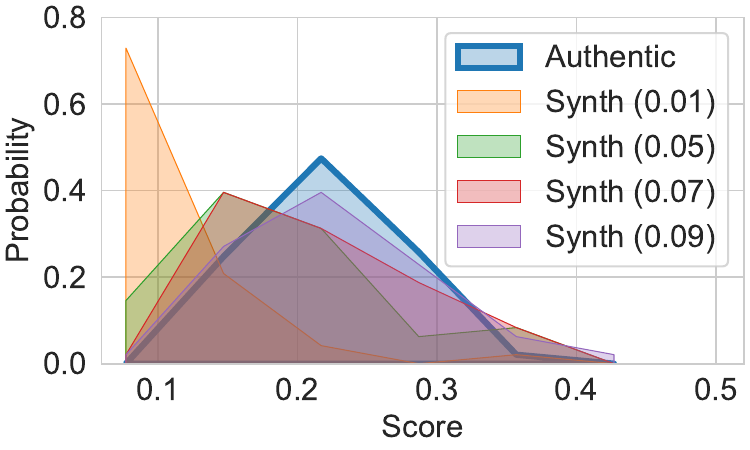}
            \caption{PMI: 361--384 hours}
            \label{fig:class16-noise-score-dist}
        \end{subfigure}\hfill
        \begin{subfigure}{0.3\linewidth}
            \includegraphics[width=\linewidth]{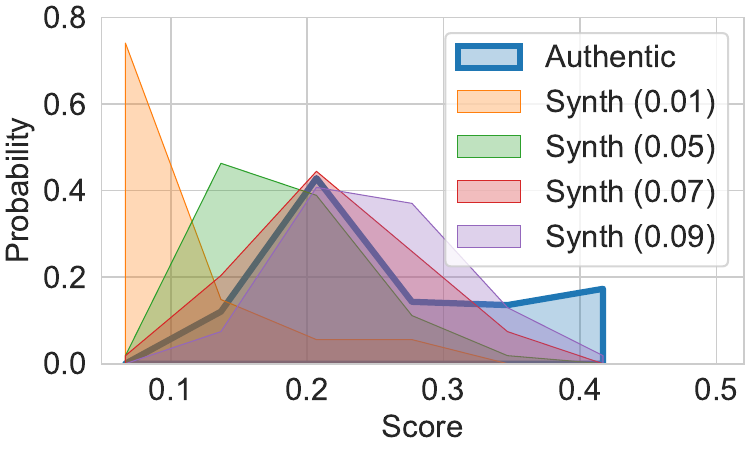}
            \caption{PMI: 385--408 hours}
            \label{fig:class17-noise-score-dist}
        \end{subfigure}\hfill
        \begin{subfigure}{0.3\linewidth}
            \includegraphics[width=\linewidth]{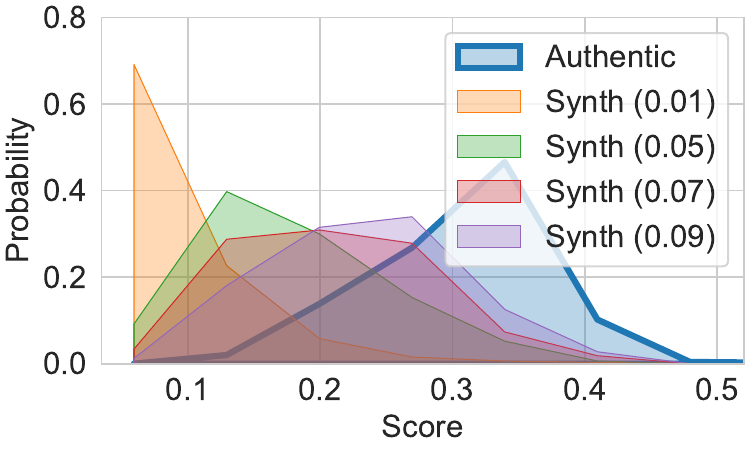}
            \caption{PMI: 409--1674 hours}
            \label{fig:class18-noise-score-dist}
        \end{subfigure}

    \caption{Same as in Fig. \ref{fig:entire-genuine-auth-synth-score-dist}, except for breaking the plots by the PMI range.}
    \label{fig:class-wise-noise-level-score-dist}
\end{figure*}